\documentclass{article}

\usepackage{algorithm}
\usepackage{nicefrac, xfrac}
\usepackage{amsmath,amssymb}
\usepackage{graphicx}
\usepackage{tabularx}
\usepackage{subcaption}
\usepackage{multirow}

\usepackage{amsmath,amsfonts,bm}

\def\eqref#1{eq. ~\ref{#1}}

\def\1{\bm{1}}

\newcommand{\normal}{\mathcal{N}}

\def\rvc{{\mathbf{c}}}

\def\rvx{{\mathbf{x}}}

\def\rmI{{\mathbf{I}}}

\DeclareMathAlphabet{\mathsfit}{\encodingdefault}{\sfdefault}{m}{sl}
\SetMathAlphabet{\mathsfit}{bold}{\encodingdefault}{\sfdefault}{bx}{n}

\newcommand{\E}{\mathbb{E}}

\usepackage[numbers]{natbib}
\usepackage[preprint]{neurips_2023}

\usepackage[utf8]{inputenc} 
\usepackage[T1]{fontenc}    
\usepackage{hyperref}       
\usepackage{url}            
\usepackage{booktabs}       
\usepackage{amsfonts}       
\usepackage{nicefrac}       
\usepackage{microtype}      
\usepackage{xcolor}         

\title{GD-VDM: Generated Depth for better Diffusion-based Video Generation}

\author{%
  Ariel Lapid \\
  Bar-Ilan University, Israel\\
  \texttt{lapida@biu.ac.il} \\
  \And
  Idan Achituve \\
  Bar-Ilan University, Israel\\
  \texttt{idan.achituve@biu.ac.il} \\
  \AND
  Lior Bracha \\
  Bar-Ilan University, Israel\\
  \texttt{brachal@biu.ac.il} \\
  \And
  Ethan Fetaya \\
  Bar-Ilan University, Israel\\
  \texttt{ethan.fetaya@biu.ac.il} \\
}

\begin{document}

\maketitle

\begin{abstract}
The field of generative models has recently witnessed significant progress, with diffusion models showing remarkable performance in image generation. In light of this success, there is a growing interest in exploring the application of diffusion models to other modalities. One such challenge is the generation of coherent videos of complex scenes, which poses several technical difficulties, such as capturing temporal dependencies and generating long, high-resolution videos. This paper proposes GD-VDM, a novel diffusion model for video generation, demonstrating promising results. GD-VDM is based on a two-phase generation process involving generating depth videos followed by a novel diffusion Vid2Vid model that generates a coherent real-world video. We evaluated GD-VDM on the Cityscapes dataset and found that it generates more diverse and complex scenes compared to natural baselines, demonstrating the efficacy of our approach. Our implementation is available at \href{https://github.com/lapid92/GD-VDM}{https://github.com/lapid92/GD-VDM}
\end{abstract}

\section{Introduction}
\label{sec:intro}

In recent years, advances in diffusion models have resulted in significant improvements to image generation capabilities, enabling the creation of high-quality and diverse images \cite{rombach2022high, ho2022cascaded}. Despite the successful application of diffusion models in image generation, their application in video generation has not yet attained the same level of success. One difficulty in video generation is that they not only need to learn the challenging image generation task but also the dynamical model on top of it. A natural extension of diffusion models to videos was first presented in \cite{ho2022video} under the name Video Diffusion Models (VDM).
When applying this model to the SUN3D dataset \cite{6751312} with a moving camera in a static scene, the model produced good-looking videos that, at least qualitatively, matched the training data. However, when generating more complex scenes, such as driving scenes having other moving objects, we often found this method to generate monotonic videos with a similar structure, e.g., a driving scene of an empty road. This presents a challenge - how to improve video generation to make our model generate more diverse and complex videos, capturing the actual training distribution while maintaining a high-quality video. \\

Here, we address this problem through a two-stage video generation process. In the first phase, we generate a video of depth images. The intuition behind this stage is that it allows us to ignore many of the fine details and textures in the video and focus on the scene composition and dynamics. In the second phase, we train a conditional video-to-video diffusion model (Vid2Vid) that generates a real-looking video conditioned on the depth video. We found that videos generated with this approach had more complex scenes and included more naturally moving objects, such as other vehicles. We named our approach Generated-Depth Video Diffusion Model or \textit{GD-VDM}. \\

\begin{figure}[!t]
  \centering
  \begin{tabular}  {m{0.0\textwidth}m{0.13\textwidth}m{0.13\textwidth}m{0.13\textwidth}m{0.13\textwidth}m{0.13\textwidth}} 
  \rotatebox{90} {\scriptsize{VDM \cite{ho2022video}}} &
    \includegraphics[width=0.15\textwidth]{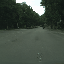} & \includegraphics[width=0.15\textwidth]{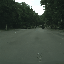} & \includegraphics[width=0.15\textwidth]{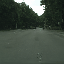} & \includegraphics[width=0.15\textwidth]{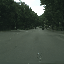} & \includegraphics[width=0.15\textwidth]{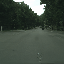} \\
    \rotatebox{90} {\scriptsize{GD-VDM}} &
    \includegraphics[width=0.15\textwidth]{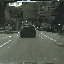} & \includegraphics[width=0.15\textwidth]{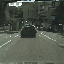} & \includegraphics[width=0.15\textwidth]{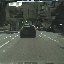} & \includegraphics[width=0.15\textwidth]{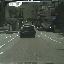} & \includegraphics[width=0.15\textwidth]{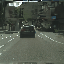} \\
    \rotatebox{90} {\scriptsize{GD-VDM} \tiny{denoised}} &
    \includegraphics[width=0.15\textwidth]{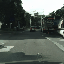} & \includegraphics[width=0.15\textwidth]{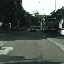} & \includegraphics[width=0.15\textwidth]{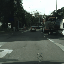} & \includegraphics[width=0.15\textwidth]{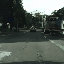} & \includegraphics[width=0.15\textwidth]{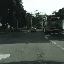} \\
    \rotatebox{90} {\scriptsize{Ground Truth}} &
    \includegraphics[width=0.15\textwidth]{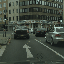} & \includegraphics[width=0.15\textwidth]{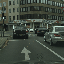} & \includegraphics[width=0.15\textwidth]{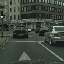} & \includegraphics[width=0.15\textwidth]{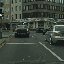} & \includegraphics[width=0.15\textwidth]{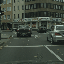} \\
  \end{tabular}
  \vspace{0.2cm}
  \caption{5-frame sequences generated from 3 methods trained on the Cityscapes dataset and a Ground Truth sequence.}
  \label{fig:1}
\end{figure}

However, one issue that arose from this hierarchical approach is some degradation in video quality compared to VDM. We hypothesize that this is due to the domain shift between training time and test time. When training the model in the second phase, we used depth videos, but when generating novel videos, we can only rely on artificially generated depth videos from the model trained in the first phase. 
Hence, to overcome this gap, we adjusted our training procedure such that the input depth videos will be more closely align with depth videos generated by our model. During training, we injected noise into the actual depth video via the forward diffusion process, followed by denoising the resultant video with our depth diffusion model. By doing so, we were able to improve the quality of the generated video. \\

To summarize this study, we make the following novel contributions:
\begin{itemize}
    \item We propose GD-VDM, a two-phased diffusion-based model capable of generating diverse, high-quality complex videos.
    \item  We introduce Vid2Vid-DM, a video-to-video translation diffusion model.
    \item We devise a simple new technique to handle domain shift.
    \item We outperform current VDM in terms of the diversity and complexity of the generated scenes.
\end{itemize}

\begin{figure}[!t]
\centering
\includegraphics[width=12cm]{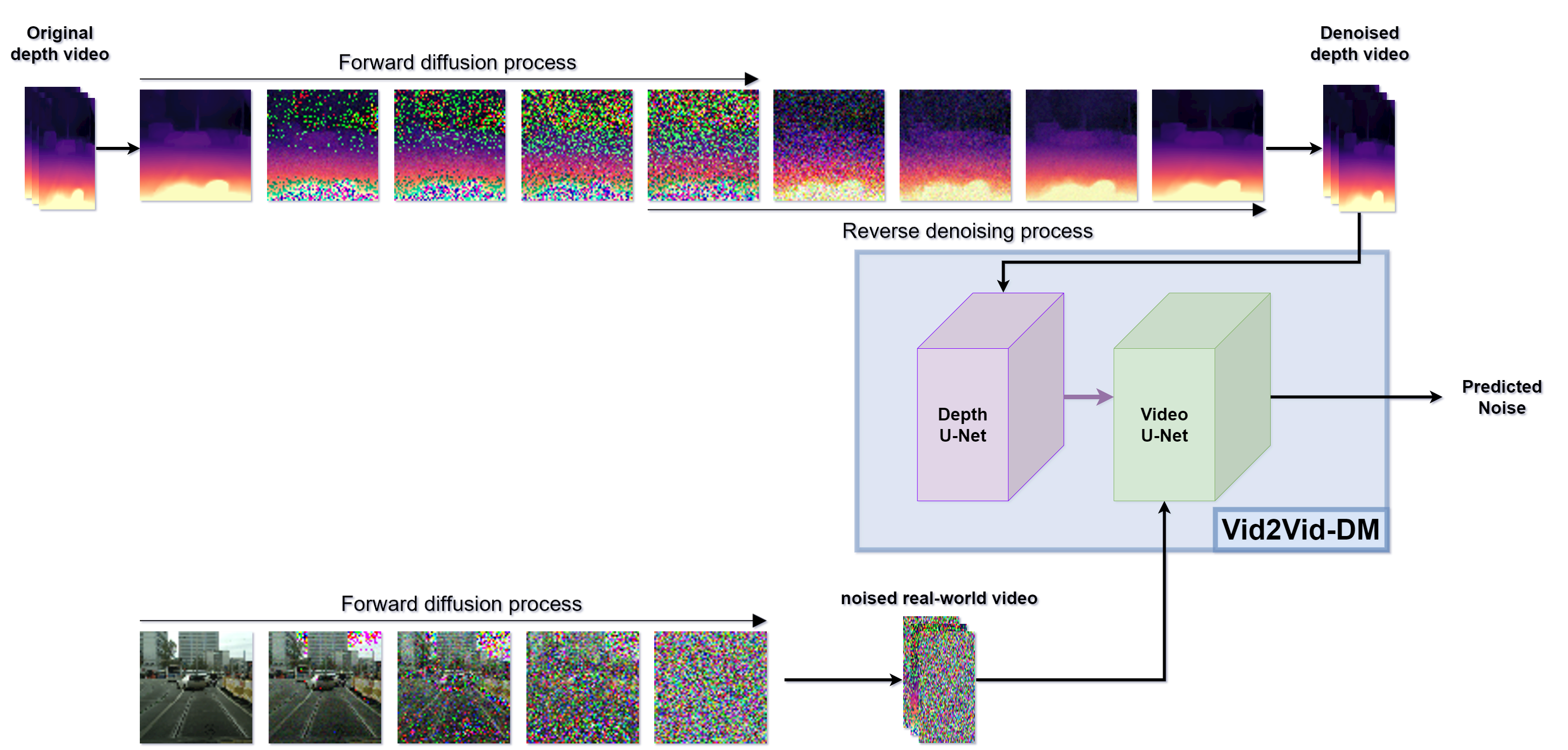}
\caption{Overview of the Vid2Vid-DM architecture - The architecture utilizes a specific approach to train the network. Initially, the architecture applies a forward noising process and then a reverse denoising process to the source depth maps to simulate inference artifacts. The resulting denoised depth sequences are fed into the Depth U-Net of the Vid2Vid-DM. The real-world video is also processed through a forward noising process, and the Video U-Net of the Vid2Vid-DM takes in the noised video and the output of the first U-Net, which is conditioned through concatenation skip connections.}
\label{fig:2}
\end{figure}

\section{Background}
\subsection[Denoising Diffusion Probabilistic Models (DDPM)]{Denoising Diffusion Probabilistic Models (DDPM) \cite{sohl2015deep, ho2020denoising}}
Assume a training data point is sampled from the data distribution we wish to model $\rvx_0 \sim p_{data}(\rvx)$. DDPMs define a discrete Markov chain with $T$ time-steps that gradually adds Gaussian noise to the data with pre-defined noise schedule $\{\beta_1, ..., \beta_t, ..., \beta_T\}$ according to $q(\rvx_{t} | \rvx_{t-1}) = \normal(\rvx_{t}| \sqrt{1 - \beta_t} \rvx_{t-1}, \beta_t \rmI)$. Thus, from Gaussian properties it follows that $q(\rvx_{t} | \rvx_{0}) = \normal(\rvx_{t}| \sqrt{\Bar{\alpha_t}} \rvx_{0}, (1 - \Bar{\alpha_t}) \rmI)$ with $\Bar{\alpha_t} = \Pi_{s=1}^{t} \alpha_s$, and $\alpha_t = 1 - \beta_t$. A sample of $\rvx_t$ can be obtained by sampling $\epsilon \sim \normal(0, \rmI)$, and using the reparemerization, $\rvx_{t} = \sqrt{\Bar{\alpha_t}} \rvx_{0} + \sqrt{1 - \Bar{\alpha_t}} \epsilon$. This transition from image to noise is known as the \textit{forward} process. Under mild conditions on the noise schedule and the number of time-steps, $q(\rvx_T)$ is approximately a standard Gaussian distribution. The goal is to train a neural network (NN) that approximates the backward transition starting from a Gaussian noise to a natural image. 
Assume that $p_{\theta}(\rvx_{t-1} | \rvx_{t})$ follows an isotropic Gaussian distribution. The mean of the distribution, $\mu_{\theta}(\rvx_t, t)$, is parameterized by $\theta$, the weights of a NN, and the covariance is the identity matrix scaled by $\beta_t$. \citet{ho2020denoising} showed that in order to learn $p_{\theta}(\rvx_{t-1} | \rvx_{t})$ we can learn to recover the added noise $\epsilon_{\theta}(\rvx_t, t)$ instead of $\mu_{\theta}(\rvx_t, t)$, and the loss function used for this purpose is reduced to $\E_{t, \rvx_0, \epsilon} ||\epsilon - \epsilon_{\theta}(\rvx_t, t)||^2$. 

\subsection[Classifier-Free Guidance]{Classifier-Free Guidance \cite{ho2022classifierfree}}
Classifier guidance \cite{dhariwal2021diffusion} aims at controlling the sample fidelity at the expense of diversity based on the gradient of a classifier imposed over a pre-trained diffusion model. Classifier-free guidance \cite{ho2022classifierfree} extends this idea by removing the dependence on a classifier. During training, 
instead of conditioning the diffusion model based on $\rvx_t$ only, it is conditioned on both $\rvx_t$ and some context $\rvc$ associated with $\rvx_0$. For instance, $\rvc$ can be the class label of $\rvx_0$ or the token $\emptyset$ for allowing the model to learn unconditional generation. The diffusion model is then trained to reconstruct the noise at each step similarly to the unconditional model. To sample from this model \citet{ho2022classifierfree} suggest using a linear combination of the conditional and unconditional noise estimates, namely $\epsilon_{\theta}(\rvx_t, \rvc, t)+\omega(\epsilon_{\theta}(\rvx_t, \rvc, t) -\epsilon_{\theta}(\rvx_t, \emptyset, t))$. Using $\omega=0$ results in a simple conditional diffusion model, while larger values of $\omega$ will lead towards a generation of a sample that depends on the context more heavily.  

\section{Method}
Our video generation model is composed of two main components. First, we use the VDM \cite{ho2022video}, which is trained on depth videos, to create a synthetic depth video. This synthetic video serves as a conditional input for the Video-to-Video Diffusion Model (Vid2Vid-DM), which generates the real-world video. \\

To overcome the domain shift challenge, we adjusted our training approach by adding noise to the depth videos through the forward diffusion process and using our depth diffusion model to denoise the resulting video. This process, shown in Fig. \ref{fig:2},  helps  simulate the inference artifacts. Our Vid2Vid-DM network is a combination of two U-Nets. The first receives the conditioning video, depth in our experiments, and produces features at various scales. The second U-Net receives the noisy video and the conditioning video features and predicts the noise. This is illustrated in Fig. \ref{fig:3} (See Appendix \ref{appendix:architecture_details} for details of this architecture.

\begin{figure}[!t]
  \centering
  \includegraphics[width=1.\textwidth]{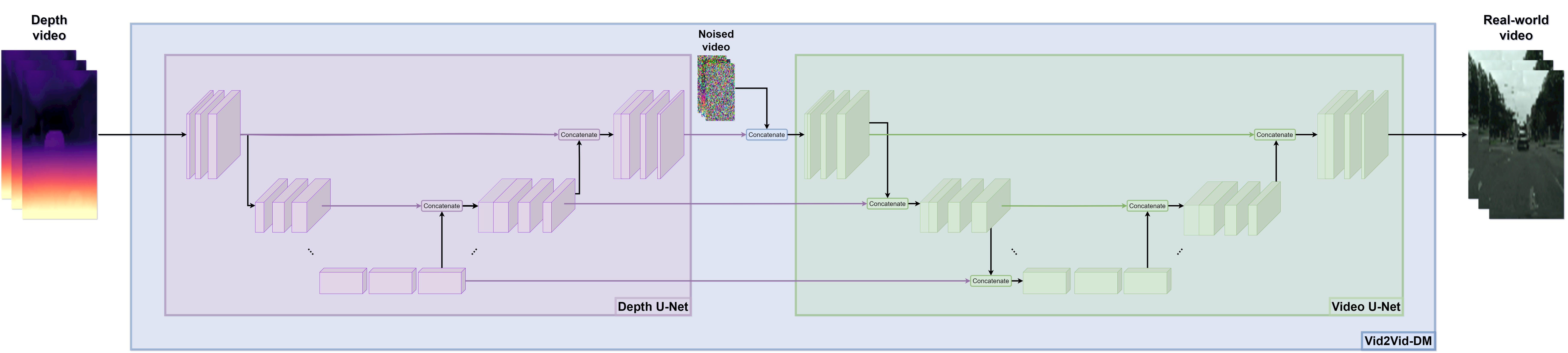}
  \caption{Overview of the vid2vid architecture}
  \label{fig:3}
\end{figure}

\subsection{Depth Generation}
As the number of available videos with depth is limited, and since we do not want our method to be dependent on the existence of additional depth information, we train with artificial depth generated from Monodepth2 \cite{godard2019digging}, an image-to-depth model. We note that the training depth videos were generated frame by frame, but we did not observe any temporal discontinuities due to this process. We also note that we cannot use this model to generate novel videos as it requires an input video. Hence, we must learn a depth diffusion model in addition to the video diffusion model. \\

We synthesize depth videos using the VDM \cite{ho2022video} framework, trained on the previously described depth videos dataset. The VDM is based on the U-Net \cite{ronneberger2015unet} network that is commonly used for image diffusion models \cite{dhariwal2021diffusion, ho2020denoising}. The U-Net architecture contains a series of downsampling layers to decrease the input's resolution, followed by an upsampling path restoring the original resolution. Furthermore, the U-Net contains concatenation skip connections between the downsampling and upsampling pass. The network used in \cite{ho2022video} consists of 3D convolutional residual blocks, each followed by a spatial-temporal attention block. \\

\subsection{Video to Video Diffusion Model}
Here we describe our Video-to-Video Diffusion Model (Vid2Vid-DM) that generates a video conditioned on another video, the depth video in our case. There are few solutions to the video-to-video translation problem in the GAN literature \cite{shi2016real, wang2018vid2vid}, but the video-to-video generation with diffusion-based models is less explored.\\

We extend the VDM to a video-conditional diffusion model by conditioning the source video via concatenation, similar to \cite{saharia2022palette, saharia2021image}. We have created a novel architecture that merges two 3D U-Net architectures. The first U-Net processes the conditioning video. It then feeds its features to the second U-Net, which also gets the noisy video and is trained to predict the noise. The first U-Net serves as a multi-scale feature extractor for the conditioning video. The second U-Net takes as input the original video, with added Gaussian noise, and at each resolution scale is concatenated with the appropriate features from the first U-Net. An illustration of the architecture is presented in Figure \ref{fig:3}.

\subsection{Generated Depth for better Diffusion-based Video Generation}
At inference, we generate a depth video using our Depth VDM model. These videos are processed through the Vid2Vid-DM to produce our generated video. However, we observed a degradation in the quality of the generated frames, which we suspected was due to a domain shift. The depth generated by our depth diffusion model is different from the original depth the Vid2Vid-DM and the depth diffusion model were trained on. To address this issue, we modified our training approach to align the depth videos more closely with those generated by our model. Specifically, we added noise to the actual depth video through the forward diffusion process, followed by denoising the resultant video with our depth diffusion model (as seen in Fig. \ref{fig:2}).

\section{Related Work}
Our approach is inspired by SB-GAN \cite{SemanticBottleneck}, a GAN-based generative model for generating images that generates images by a similar two-phase approach (with semantic segmentation instead of depth). GD-VDM shares the same fundamental concept; however, it differs from it in two key aspects. First, our approach is tailored for diffusion models. And second, we propose a natural way to address the domain shift observed between training and generation time. We also note the concatenating two U-Nets was previously proposed in \cite{xia2017wnet} as W-Net.\\

\textbf{Diffusion models}. In recent years, diffusion models and score-based generative models \cite{sohl2015deep, ho2020denoising, song2021scorebased} have shown promising results in producing high-quality samples and became the leading approach for image generation. For instance, diffusion models were successfully applied to image synthesis \cite{rombach2022high, dhariwal2021diffusion}, image-to-image translation \cite{saharia2022palette, saharia2022image, whang2022deblurring}, text-to-image translation \cite{jiang2022text2human, saharia2022photorealistic, ramesh2022hierarchical}, and audio generation \cite{chen2020wavegrad, kong2020diffwave}. In our work, we use the DDPM model \cite{ho2020denoising}, which is trained to denoise samples corrupted by varying levels of Gaussian noise. Samples are generated using a Markov chain, which progressively denoises white noise into an image either via Langevin dynamics \cite{song2020improved} or by reversing a forward diffusion process.\\

\textbf{Video generation}.
Until recently, GANs have been the most popular method for video generation. Video-GAN (VGAN) \cite{vondrick2016generating} was among the first GAN-based models to generate videos. VGAN uses a structured approach of enforcing a static background with a moving foreground. Temporal GAN (TGAN) \cite{saito2017temporal} is another GAN-based model consisting of a temporal generator and an image generator. Motion Content GAN (MoCoGAN) \cite{tulyakov2017mocogan} models videos as content space and motion space and employs both an image discriminator and a video discriminator to ensure that each frame looks natural. Dual video discriminator GAN (DVD-GAN) \cite{clark2019adversarial} expands BigGAN \cite{brock2019large} to the video domain using two discriminators, similar to MoCoGAN.
StyleGAN-V \cite{skorokhodov2022stylegan} based on StyleGAN2 \cite{karras2020analyzing}, uses a global latent code to control the content of the entire video, similar to MoCoGAN.\\

Diffusion models have made significant progress in recent years and are now being applied to video generation. The VDM \cite{ho2022video} is a recent approach that applies diffusion models to video generation tasks. VDM utilizes a 3D U-Net  \cite{cciccek20163d} architecture to handle the temporal domain in videos. When tested on complex driving scenes with moving vehicles, most of the generated videos were of good quality but mostly displayed simple, empty roads. 
Several recent works extended and improved VDM \cite{ho2022imagen,brooks2022generating,blattmann2023align}. However, they modified the architecture and training process, which is orthogonal to our contribution and can be easily combined. \\

\textbf{Video-to-Video translation}.
While the image-to-image translation problem has been widely studied and has numerous research works dedicated to it \cite{isola2017image, wang2018high, zhu2017toward, zhao2022egsde, wolleb2022swiss, saharia2022palette}, and conditional video generation is gaining traction \cite{harvey2022flexible, ho2022imagen, singer2022make, zhou2022magicvideo}, the video-to-video synthesis problem has not received as much attention in the literature \cite{wang2018vid2vid, wang2019few}. Most of the current solutions for this problem rely on GAN-based models. However, there is still significant potential for advancing and developing more efficient models for video-to-video synthesis.

\begin{figure}[!t]
  \centering
  \begin{tabular}  {m{0.0\textwidth}m{0.13\textwidth}m{0.13\textwidth}m{0.13\textwidth}m{0.13\textwidth}m{0.13\textwidth}}
  \rotatebox{90} {\scriptsize{VDM\cite{ho2022video}}} &
    \includegraphics[width=0.15\textwidth]{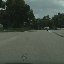} & \includegraphics[width=0.15\textwidth]{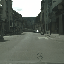} & \includegraphics[width=0.15\textwidth]{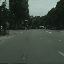} & \includegraphics[width=0.15\textwidth]{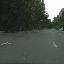} & \includegraphics[width=0.15\textwidth]{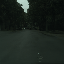} \\
    \rotatebox{90} {\scriptsize{GD-VDM}} &
    \includegraphics[width=0.15\textwidth]{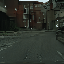} & \includegraphics[width=0.15\textwidth]{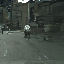} & \includegraphics[width=0.15\textwidth]{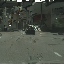} & \includegraphics[width=0.15\textwidth]{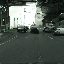} & \includegraphics[width=0.15\textwidth]{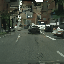} \\
    \rotatebox{90} {\scriptsize{GD-VDM} \tiny{denoised}} &
    \includegraphics[width=0.15\textwidth]{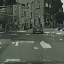} & \includegraphics[width=0.15\textwidth]{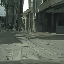} & \includegraphics[width=0.15\textwidth]{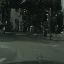} & \includegraphics[width=0.15\textwidth]{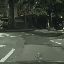} & \includegraphics[width=0.15\textwidth]{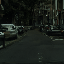} \\
    \rotatebox{90} {\scriptsize{Ground Truth}} &
    \includegraphics[width=0.15\textwidth]{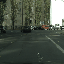} & \includegraphics[width=0.15\textwidth]{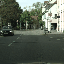} & \includegraphics[width=0.15\textwidth]{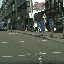} & \includegraphics[width=0.15\textwidth]{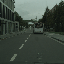} & \includegraphics[width=0.15\textwidth]{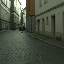} \\
  \end{tabular}
  \vspace{0.2cm}
  \caption{Randomly selected frames from randomly generated videos, ours have more diverse and complex driving scenes. We provide more examples in Appendix \ref{appendix:cityscapes_examples} .}
  \label{fig:4}
\end{figure}

\section{Experiments}
We evaluated the performance of GD-VDM, both with and without the denoised mechanism of the depth videos during training, against two baselines - a diffusion model and a GAN-based model. Our evaluation included both human evaluation and the Fréchet Video Distance (FVD) metric. However, we observed that the FVD score did not correlate strongly with the perceived video quality, a concern raised by other studies as well (e.g., \cite{brooks2022generating}). Therefore, we primarily relied on human evaluation. Nevertheless, for completeness, we provide the FVD results in \ref{appendix:fvd}.

\subsection{Training details}
In order to handle the computational requirements of training video models, we utilized a training dataset composed of videos consisting of 10 frames with a 64X64 spatial resolution. Our training approach consists of two stages. We trained the VDM on depth videos in the first stage. Then, we proceeded to the Vid2Vid Diffusion Model training. In the second stage, we use the weights of the depth video diffusion model obtained in the first phase to initialize the depth U-Net. However, we do not keep the weights fixed and update them during the training process.

\begin{figure}[!t]
  \centering
  \begin{tabular}  {m{0.0\textwidth}m{0.13\textwidth}m{0.13\textwidth}m{0.13\textwidth}m{0.13\textwidth}m{0.13\textwidth}}
  \rotatebox{90} {\scriptsize{Generated Depth}} &
    \includegraphics[width=0.15\textwidth]{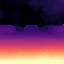} & \includegraphics[width=0.15\textwidth]{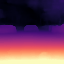} & \includegraphics[width=0.15\textwidth]{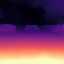} & \includegraphics[width=0.15\textwidth]{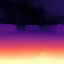} & \includegraphics[width=0.15\textwidth]{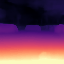} \\
    \rotatebox{90} {\scriptsize{GD-VDM}} &
    \includegraphics[width=0.15\textwidth]{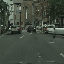} & \includegraphics[width=0.15\textwidth]{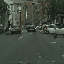} & \includegraphics[width=0.15\textwidth]{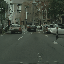} & \includegraphics[width=0.15\textwidth]{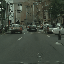} & \includegraphics[width=0.15\textwidth]{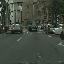} \\
    \rotatebox{90} {\scriptsize{GD-VDM} \tiny{(denoised}} &
    \includegraphics[width=0.15\textwidth]{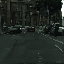} & \includegraphics[width=0.15\textwidth]{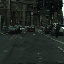} & \includegraphics[width=0.15\textwidth]{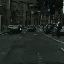} & \includegraphics[width=0.15\textwidth]{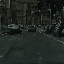} & \includegraphics[width=0.15\textwidth]{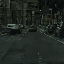} \\
  \end{tabular}
  \vspace{0.2cm}
  \caption{Generated depth and the synthesized video produced conditioned on him.}
  \label{fig:5}
\end{figure}

\subsection{Datasets}
Our proposed approach was evaluated using the Cityscapes dataset \cite{cordts2016cityscapes}, which comprises of 30-frame scene videos of streets captured in various German cities, with a resolution of 2048 × 1024. The training set consists of 2975 videos, each with 30 frames, while the validation set comprises 500 videos, and the test set contains 1525 videos, each with 30 frames. Only a subset of images in the videos contains ground truth depth maps. To obtain the depth map videos, we leveraged recent advances in neural network-based depth estimation on monocular videos \cite{zhou2017unsupervised, zhan2018unsupervised, godard2019digging}. Specifically,  we adopt Monodepth2 \cite{godard2019digging}, trained on Cityscapes, to generate a dataset of depth videos to train our depth diffusion model. We then created a second depth dataset, named the denoised-depth dataset, which we used to train our Vid2Vid-DM model. We achieved this by introducing noise to the depth video through forward diffusion and denoising the resulting video using our depth diffusion model, as illustrated in Fig. \ref{fig:2}.

We also conducted experiments on the SUN3D dataset \cite{6751312}, which contains 415 sequences captured in 254 distinct areas across 41 buildings.  We saw in our experiments  that all methods were able to generate visually pleasing videos that qualitatively matched the training data. We attribute the good performance of the VDM on this dataset to the fact that the videos have simple scenes with a moving camera in a static background.
We provide samples from models trained on the SUN3D dataset in \ref{appendix:sun3d_examples}.

\subsection{Baselines}
We utilized two models for comparison purposes: \textbf{Video Diffusion Model} \cite{ho2022video} and \textbf{MoCoGAN} \cite{tulyakov2017mocogan}. The VDM is a 3D U-Net-based video diffusion model capable of generating high-definition videos with strong temporal consistency. MoCoGAN \cite{tulyakov2017mocogan}, on the other hand, is a GAN-based model that can learn motion and content in an unsupervised manner and divides the latent image space into content and motion spaces. Both of these models were trained on the Cityscapes video dataset.

\begin{table*}[!t]
\centering
\scalebox{.85}{
    \begin{tabular}{l c cc cc cc c}
    \toprule
    && Diversity && Quality && Richness && Cars\\
    \midrule
    Ground Truth && 6.50 $\pm$ 0.44 && 5.59 $\pm$ 0.10 && 4.71 $\pm$ 0.14 && 20\%\\
    \hline
    MoCoGAN \cite{tulyakov2017mocogan} && 3.59 $\pm$ 0.13 && 3.30 $\pm$ 0.03 && 3.12 $\pm$ 0.09 && 4\% \\
    Video Diffusion Model \cite{ho2022video} && 4.17 $\pm$ 0.16 && \textbf{4.31} $\pm$ 0.05 && 3.83 $\pm$ 0.11 && 7\% \\
    \midrule
    GD-VDM (Ours) && 4.47 $\pm$ 0.14 && 3.78 $\pm$ 0.04 && 3.97 $\pm$ 0.10 && \textbf{15\%}\\
    GD-VDM denoised (Ours) && \textbf{4.73} $\pm$ 0.17 && 3.91 $\pm$ 0.04 && \textbf{4.13} $\pm$ 0.12 && \textbf{15\%}\\
    \bottomrule
    \end{tabular}
}
\vspace{0.2cm}
\caption{Human evaluation of generated videos compared across different parameter values.}
\label{tab:user_study}
\end{table*}

\subsection{Human evaluation and analysis}
To assess the visual quality, diversity, and complexity of the synthesized videos, we conducted a human evaluation using Amazon Mechanical Turk. Our user study involved the creation of 100 videos for each approach, including baselines and the Cityscapes test data, each of 10 frames. In the evaluation, participants on AMT were shown a set of 10 videos generated using a specific approach, each set reviewed by three annotators. The purpose of simultaneously presenting 10 videos of the same method was to assess the level of diversity among  the generated videos.  We also provided two reference videos - one "good" video from the Cityscapes test data and one "bad" video with an all-black background - to help participants calibrate their responses. \\

We requested that the participants rate the diversity of the series of videos and the quality and richness of each specific video. Finally, participants were also asked whether or not a moving car was present in the video. We considered a moving car present if there was a consensus among all 3 participants.
Table \ref{tab:user_study} shows that our approach has better diversity and scene complexity compared to the baselines. Our models also have double the number of moving vehicles in the generated videos compared to the baselines. However, regarding video quality, VDM receives better results than our method. This is unsurprising, as it mainly generated simple and easy scenes, as seen in Fig. \ref{fig:4} and in Fig. \ref{fig:1}. We also see that training with the denoised depth, denoted as {Ours} {(denoised)}, improves results in all metrics. Figure \ref{fig:5} displays the difference between the generated videos between the model that underwent training with denoised depth and the model trained on real-depth videos. We provide further information about the user study in \ref{appendix:human_evaluation}.

\section{Conclusion}
We present GD-VDM, a two-phased diffusion model for video generation that shows promising results. We propose a two-phase generation approach: First, we generate the depth which defines the scene layout but without many of the small details, and then use it to generate the fully detailed video. By utilizing this approach, we demonstrate that diffusion models can learn more complex video scenes. Additionally, we introduce a diffusion Vid2Vid model that generates coherent and realistic videos based on the conditioned depth video.
We evaluated our approach against VDM using both human evaluation and the FVD metric. As a result, our model produced more diverse videos and presented complex scenes. These findings highlight the potential of diffusion models in advancing video generation tasks, and we hope that our work inspires further research and development in this exciting direction.
\bibliography{egbib}

\newpage\hbox{}\thispagestyle{empty}

\appendix

\section{Architecture details}
\label{appendix:architecture_details}
\subsection{Depth Video Diffusion Model Hyperparameters}

\begin{table}[ht]
\centering
\footnotesize
\begin{tabular}
{|c|c|}
\hline
\textbf{Parameter} & \textbf{Value} \\
\hline
Base channels &  64 \\
\hline
Channel multipliers & 1, 2, 4, 8 \\
\hline
Blocks per resolution & 2 \\
\hline
Attention head dimension & 32 \\
\hline
Frame resolution & 64x64 \\
\hline
Number of frames & 10 \\
\hline
Input channels & 3 \\
\hline
Output channels & 3 \\
\hline
Diffusion timesteps & 1000 \\
\hline
Diffusion noise schedule & cosine \\
\hline
Prediction target & $\epsilon$ \\
\hline
\end{tabular}
\end{table}

\subsection{Video to Video Diffusion Model Hyperparameters}
\begin{table}[ht]
\centering
\footnotesize
\begin{tabular}
{|c|c|}
\hline
\textbf{Parameter} & \textbf{Value} \\
\hline
Base channels &  64 \\
\hline
Channel multipliers & 1, 2, 4, 8 \\
\hline
Blocks per resolution & 2 \\
\hline
Attention head dimension & 32 \\
\hline
Frame resolution & 64x64 \\
\hline
Number of frames & 10 \\
\hline
Input channels & 3 \\
\hline
Output channels & 3 \\
\hline
Diffusion timesteps & 1000 \\
\hline
Diffusion noise schedule & cosine \\
\hline
Prediction target & $\epsilon$ \\
\hline
Condition dropout probability & 0.2 \\
\hline
Guidance weight & 1.4 \\
\hline
\end{tabular}
\end{table}

\section{Fréchet Video Distance (FVD) results}
\label{appendix:fvd}
Our evaluation of the generated videos included the Fréchet Video Distance (FVD) \cite{unterthiner2019accurate}, which compares the realism of samples by computing a 2-Wasserstein distance between the distribution of the ground truth videos and the distribution of videos generated by the model. In general, this metric relies on a pre-trained I3D network to obtain low-dimensional feature representations, whose distributions are then utilized in the computation of the FVD. \\
We outperform the VDM and the MoCoGAN in terms of the FVD score. Yet, our denoised approach yields inferior results compared to our standard method, contrary to our expectations. We observed that the FVD score did not strongly correlate with perceived video quality. This concern was also noted in \cite{brooks2022generating}. In the paper, we mainly rely on human evaluation to assess the performance of our method. Therefore we included the FVD results here for completeness.

\begin{table}[ht]
\centering
\begin{tabular}{|c|c|}
\hline
\textbf{Method} & \textbf{FVD $\downarrow$}\\
\hline
\textbf{Video Diffusion Model \cite{ho2022video}} & 224 \\
\hline
\textbf{MoCoGAN \cite{tulyakov2017mocogan}} & 588 \\
\hline
\textbf{Ours} & \textbf{127} \\
\hline
\textbf{Ours} {\footnotesize (denoised training)} & 149 \\
\hline
\end{tabular}
\vspace{0.5cm}
\caption{ \label{tab:fvd_compare} Comparison of FVD scores on the Cityscapes \cite{cordts2016cityscapes} dataset.}
\end{table}

\newpage

\section{Video samples}
\subsection{SUN3D}
\label{appendix:sun3d_examples}
After training our networks on the SUN3D dataset \cite{6751312}, we observed that the Video Diffusion Model \cite{ho2022video} produced videos with impressive visual quality. We believe this outcome is due to the dataset's scenes being relatively simple, consisting of a moving camera and a static background. Figure \ref{fig:6} and Figure \ref{fig:7} show generated examples from the methods trained on the dataset.

\newpage

\begin{table}[htb!]
\centering
\begin{tabular}
{cm{0.13\textwidth}m{0.13\textwidth} m{0.13\textwidth}m{0.13\textwidth}m{0.13\textwidth}}
  \multirow{1}{*}[-8ex]{\rotatebox{90} {VDM \cite{ho2022video}}} & \raisebox{-\totalheight}{\includegraphics[width=2cm]{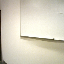}} & \raisebox{-\totalheight}{\includegraphics[width=2cm]{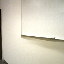}}& \raisebox{-\totalheight}{\includegraphics[width=2cm]{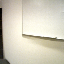}} & \raisebox{-\totalheight}{\includegraphics[width=2cm]{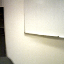}}& \raisebox{-\totalheight}{\includegraphics[width=2cm]{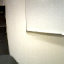}}  \\
& \raisebox{-\totalheight}{\includegraphics[width=2cm]{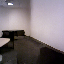}} & \raisebox{-\totalheight}{\includegraphics[width=2cm]{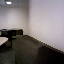}}& \raisebox{-\totalheight}{\includegraphics[width=2cm]{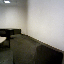}} & \raisebox{-\totalheight}{\includegraphics[width=2cm]{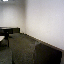}}& \raisebox{-\totalheight}{\includegraphics[width=2cm]{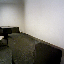}}  \\

  \multirow{1}{*}[-8ex]{\rotatebox{90} {Ours}} & \raisebox{-\totalheight}{\includegraphics[width=2cm]{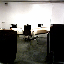}} & \raisebox{-\totalheight}{\includegraphics[width=2cm]{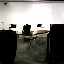}}& \raisebox{-\totalheight}{\includegraphics[width=2cm]{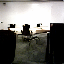}} & \raisebox{-\totalheight}{\includegraphics[width=2cm]{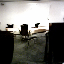}}& \raisebox{-\totalheight}{\includegraphics[width=2cm]{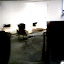}}  \\
  & \raisebox{-\totalheight}{\includegraphics[width=2cm]{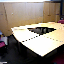}} & \raisebox{-\totalheight}{\includegraphics[width=2cm]{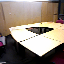}}& \raisebox{-\totalheight}{\includegraphics[width=2cm]{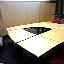}} & \raisebox{-\totalheight}{\includegraphics[width=2cm]{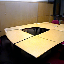}}& \raisebox{-\totalheight}{\includegraphics[width=2cm]{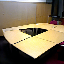}}  \\
  
  \multirow{1}{*}[-8ex]{\rotatebox{90} {Ours \tiny{(denoised}}} & \raisebox{-\totalheight}{\includegraphics[width=2cm]{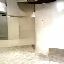}} & \raisebox{-\totalheight}{\includegraphics[width=2cm]{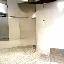}}& \raisebox{-\totalheight}{\includegraphics[width=2cm]{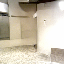}} & \raisebox{-\totalheight}{\includegraphics[width=2cm]{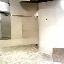}}& \raisebox{-\totalheight}{\includegraphics[width=2cm]{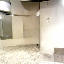}}  \\
& \raisebox{-\totalheight}{\includegraphics[width=2cm]{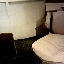}} & \raisebox{-\totalheight}{\includegraphics[width=2cm]{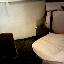}}& \raisebox{-\totalheight}{\includegraphics[width=2cm]{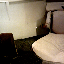}} & \raisebox{-\totalheight}{\includegraphics[width=2cm]{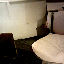}}& \raisebox{-\totalheight}{\includegraphics[width=2cm]{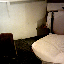}}  \\

  \multirow{1}{*}[-8ex]{\rotatebox{90} {Ground Truth}} & \raisebox{-\totalheight}{\includegraphics[width=2cm]{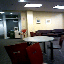}} & \raisebox{-\totalheight}{\includegraphics[width=2cm]{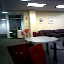}}& \raisebox{-\totalheight}{\includegraphics[width=2cm]{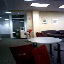}} & \raisebox{-\totalheight}{\includegraphics[width=2cm]{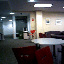}}& \raisebox{-\totalheight}{\includegraphics[width=2cm]{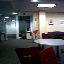}}  \\
  & \raisebox{-\totalheight}{\includegraphics[width=2cm]{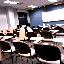}} & \raisebox{-\totalheight}{\includegraphics[width=2cm]{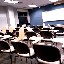}}& \raisebox{-\totalheight}{\includegraphics[width=2cm]{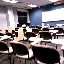}} & \raisebox{-\totalheight}{\includegraphics[width=2cm]{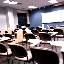}}& \raisebox{-\totalheight}{\includegraphics[width=2cm]{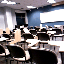}}  \\
  \end{tabular}
  \vspace{0.5cm}
  \captionof{figure}{Randomly generated videos. 5-frame sequences from 3 methods trained on the SUN3D dataset. 2 sequences of the Ground Truth dataset are provided.}
  \label{fig:6}
\end{table}

\newpage

\begin{table}[htb!]
\centering
\begin{tabular}
{cm{0.13\textwidth}m{0.13\textwidth} m{0.13\textwidth}m{0.13\textwidth}m{0.13\textwidth}}
  \multirow{1}{*}[-8ex]{\rotatebox{90} {VDM \cite{ho2022video}}} & \raisebox{-\totalheight}{\includegraphics[width=2cm]{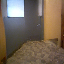}} & \raisebox{-\totalheight}{\includegraphics[width=2cm]{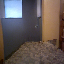}}& \raisebox{-\totalheight}{\includegraphics[width=2cm]{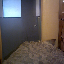}} & \raisebox{-\totalheight}{\includegraphics[width=2cm]{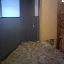}}& \raisebox{-\totalheight}{\includegraphics[width=2cm]{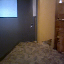}}  \\
  
& \raisebox{-\totalheight}{\includegraphics[width=2cm]{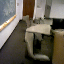}} & \raisebox{-\totalheight}{\includegraphics[width=2cm]{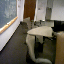}}& \raisebox{-\totalheight}{\includegraphics[width=2cm]{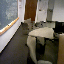}} & \raisebox{-\totalheight}{\includegraphics[width=2cm]{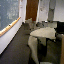}}& \raisebox{-\totalheight}{\includegraphics[width=2cm]{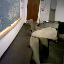}}  \\

  \multirow{1}{*}[-8ex]{\rotatebox{90} {Ours}} & \raisebox{-\totalheight}{\includegraphics[width=2cm]{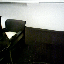}} & \raisebox{-\totalheight}{\includegraphics[width=2cm]{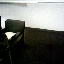}}& \raisebox{-\totalheight}{\includegraphics[width=2cm]{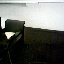}} & \raisebox{-\totalheight}{\includegraphics[width=2cm]{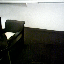}}& \raisebox{-\totalheight}{\includegraphics[width=2cm]{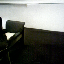}}  \\
  & \raisebox{-\totalheight}{\includegraphics[width=2cm]{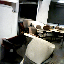}} & \raisebox{-\totalheight}{\includegraphics[width=2cm]{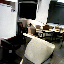}}& \raisebox{-\totalheight}{\includegraphics[width=2cm]{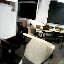}} & \raisebox{-\totalheight}{\includegraphics[width=2cm]{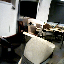}}& \raisebox{-\totalheight}{\includegraphics[width=2cm]{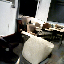}}  \\
  
  \multirow{1}{*}[-8ex]{\rotatebox{90} {Ours \tiny{(denoised}}} & \raisebox{-\totalheight}{\includegraphics[width=2cm]{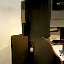}} & \raisebox{-\totalheight}{\includegraphics[width=2cm]{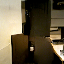}}& \raisebox{-\totalheight}{\includegraphics[width=2cm]{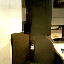}} & \raisebox{-\totalheight}{\includegraphics[width=2cm]{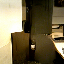}}& \raisebox{-\totalheight}{\includegraphics[width=2cm]{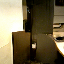}}  \\
& \raisebox{-\totalheight}{\includegraphics[width=2cm]{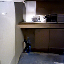}} & \raisebox{-\totalheight}{\includegraphics[width=2cm]{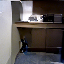}}& \raisebox{-\totalheight}{\includegraphics[width=2cm]{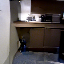}} & \raisebox{-\totalheight}{\includegraphics[width=2cm]{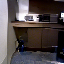}}& \raisebox{-\totalheight}{\includegraphics[width=2cm]{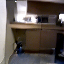}}  \\

  \multirow{1}{*}[-8ex]{\rotatebox{90} {Ground Truth}} & \raisebox{-\totalheight}{\includegraphics[width=2cm]{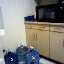}} & \raisebox{-\totalheight}{\includegraphics[width=2cm]{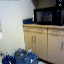}}& \raisebox{-\totalheight}{\includegraphics[width=2cm]{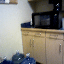}} & \raisebox{-\totalheight}{\includegraphics[width=2cm]{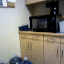}}& \raisebox{-\totalheight}{\includegraphics[width=2cm]{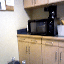}}  \\
  & \raisebox{-\totalheight}{\includegraphics[width=2cm]{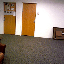}} & \raisebox{-\totalheight}{\includegraphics[width=2cm]{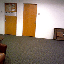}}& \raisebox{-\totalheight}{\includegraphics[width=2cm]{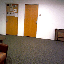}} & \raisebox{-\totalheight}{\includegraphics[width=2cm]{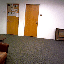}}& \raisebox{-\totalheight}{\includegraphics[width=2cm]{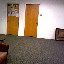}}  \\
  \end{tabular}
  \vspace{0.5cm}
  \captionof{figure}{More randomly generated videos based on the SUN3D dataset.}
  \label{fig:7}
\end{table}

\newpage

\subsection{Cityscapes}
\label{appendix:cityscapes_examples}
Figures \ref{fig:8}, \ref{fig:9} contain examples from models trained on Cityscapes \cite{cordts2016cityscapes}.

\begin{table}[htb!]
\centering
\begin{tabular}
{cm{0.13\textwidth}m{0.13\textwidth} m{0.13\textwidth}m{0.13\textwidth}m{0.13\textwidth}}
  \multirow{1}{*}[-8ex]{\rotatebox{90} {VDM \cite{ho2022video}}} & \raisebox{-\totalheight}{\includegraphics[width=2cm]{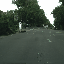}} & \raisebox{-\totalheight}{\includegraphics[width=2cm]{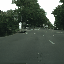}}& \raisebox{-\totalheight}{\includegraphics[width=2cm]{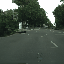}} & \raisebox{-\totalheight}{\includegraphics[width=2cm]{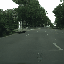}}& \raisebox{-\totalheight}{\includegraphics[width=2cm]{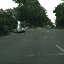}}  \\
& \raisebox{-\totalheight}{\includegraphics[width=2cm]{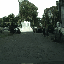}} & \raisebox{-\totalheight}{\includegraphics[width=2cm]{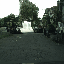}}& \raisebox{-\totalheight}{\includegraphics[width=2cm]{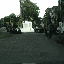}} & \raisebox{-\totalheight}{\includegraphics[width=2cm]{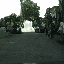}}& \raisebox{-\totalheight}{\includegraphics[width=2cm]{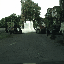}}  \\

  \multirow{1}{*}[-8ex]{\rotatebox{90} {Ours}} & \raisebox{-\totalheight}{\includegraphics[width=2cm]{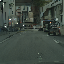}} & \raisebox{-\totalheight}{\includegraphics[width=2cm]{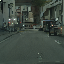}}& \raisebox{-\totalheight}{\includegraphics[width=2cm]{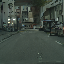}} & \raisebox{-\totalheight}{\includegraphics[width=2cm]{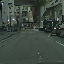}}& \raisebox{-\totalheight}{\includegraphics[width=2cm]{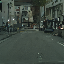}}  \\
  & \raisebox{-\totalheight}{\includegraphics[width=2cm]{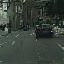}} & \raisebox{-\totalheight}{\includegraphics[width=2cm]{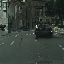}}& \raisebox{-\totalheight}{\includegraphics[width=2cm]{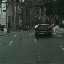}} & \raisebox{-\totalheight}{\includegraphics[width=2cm]{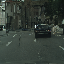}}& \raisebox{-\totalheight}{\includegraphics[width=2cm]{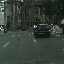}}  \\
  
  \multirow{1}{*}[-8ex]{\rotatebox{90} {Ours \tiny{(denoised}}} & \raisebox{-\totalheight}{\includegraphics[width=2cm]{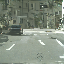}} & \raisebox{-\totalheight}{\includegraphics[width=2cm]{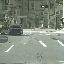}}& \raisebox{-\totalheight}{\includegraphics[width=2cm]{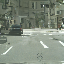}} & \raisebox{-\totalheight}{\includegraphics[width=2cm]{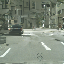}}& \raisebox{-\totalheight}{\includegraphics[width=2cm]{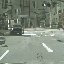}}  \\
& \raisebox{-\totalheight}{\includegraphics[width=2cm]{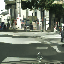}} & \raisebox{-\totalheight}{\includegraphics[width=2cm]{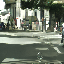}}& \raisebox{-\totalheight}{\includegraphics[width=2cm]{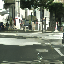}} & \raisebox{-\totalheight}{\includegraphics[width=2cm]{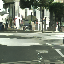}}& \raisebox{-\totalheight}{\includegraphics[width=2cm]{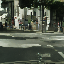}}  \\

  \multirow{1}{*}[-8ex]{\rotatebox{90} {Ground Truth}} & \raisebox{-\totalheight}{\includegraphics[width=2cm]{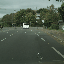}} & \raisebox{-\totalheight}{\includegraphics[width=2cm]{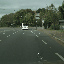}} & \raisebox{-\totalheight}{\includegraphics[width=2cm]{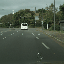}} & \raisebox{-\totalheight}{\includegraphics[width=2cm]{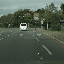}} & \raisebox{-\totalheight}{\includegraphics[width=2cm]{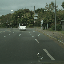}}\\
  & \raisebox{-\totalheight}{\includegraphics[width=2cm]{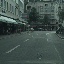}} & \raisebox{-\totalheight}{\includegraphics[width=2cm]{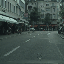}} & \raisebox{-\totalheight}{\includegraphics[width=2cm]{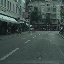}} & \raisebox{-\totalheight}{\includegraphics[width=2cm]{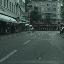}} & \raisebox{-\totalheight}{\includegraphics[width=2cm]{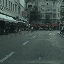}}\\
  \end{tabular}
  \vspace{0.5cm}
  \captionof{figure}{Randomly generated videos. 5-frame sequences from 3 methods trained on the Cityscapes dataset. 2 sequences of the Ground Truth dataset are provided.}
  \label{fig:8}
\end{table}

\newpage

\begin{table}[htb!]
\centering
\begin{tabular}
{cm{0.13\textwidth}m{0.13\textwidth} m{0.13\textwidth}m{0.13\textwidth}m{0.13\textwidth}}
  \multirow{1}{*}[-8ex]{\rotatebox{90} {VDM \cite{ho2022video}}} & \raisebox{-\totalheight}{\includegraphics[width=2cm]{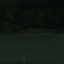}} & \raisebox{-\totalheight}{\includegraphics[width=2cm]{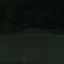}}& \raisebox{-\totalheight}{\includegraphics[width=2cm]{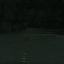}} & \raisebox{-\totalheight}{\includegraphics[width=2cm]{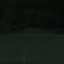}}& \raisebox{-\totalheight}{\includegraphics[width=2cm]{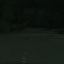}}  \\
  
& \raisebox{-\totalheight}{\includegraphics[width=2cm]{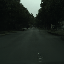}} & \raisebox{-\totalheight}{\includegraphics[width=2cm]{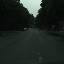}}& \raisebox{-\totalheight}{\includegraphics[width=2cm]{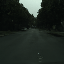}} & \raisebox{-\totalheight}{\includegraphics[width=2cm]{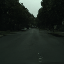}}& \raisebox{-\totalheight}{\includegraphics[width=2cm]{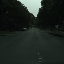}}  \\

  \multirow{1}{*}[-8ex]{\rotatebox{90} {Ours}} & \raisebox{-\totalheight}{\includegraphics[width=2cm]{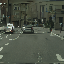}} & \raisebox{-\totalheight}{\includegraphics[width=2cm]{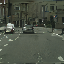}}& \raisebox{-\totalheight}{\includegraphics[width=2cm]{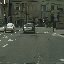}} & \raisebox{-\totalheight}{\includegraphics[width=2cm]{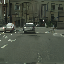}}& \raisebox{-\totalheight}{\includegraphics[width=2cm]{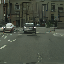}}  \\
  & \raisebox{-\totalheight}{\includegraphics[width=2cm]{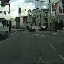}} & \raisebox{-\totalheight}{\includegraphics[width=2cm]{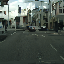}}& \raisebox{-\totalheight}{\includegraphics[width=2cm]{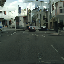}} & \raisebox{-\totalheight}{\includegraphics[width=2cm]{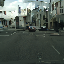}}& \raisebox{-\totalheight}{\includegraphics[width=2cm]{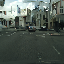}}  \\
  
  \multirow{1}{*}[-8ex]{\rotatebox{90} {Ours \tiny{(denoised}}} & \raisebox{-\totalheight}{\includegraphics[width=2cm]{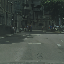}} & \raisebox{-\totalheight}{\includegraphics[width=2cm]{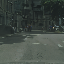}}& \raisebox{-\totalheight}{\includegraphics[width=2cm]{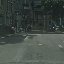}} & \raisebox{-\totalheight}{\includegraphics[width=2cm]{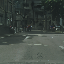}}& \raisebox{-\totalheight}{\includegraphics[width=2cm]{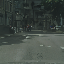}}  \\
& \raisebox{-\totalheight}{\includegraphics[width=2cm]{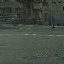}} & \raisebox{-\totalheight}{\includegraphics[width=2cm]{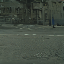}}& \raisebox{-\totalheight}{\includegraphics[width=2cm]{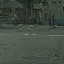}} & \raisebox{-\totalheight}{\includegraphics[width=2cm]{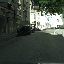}}& \raisebox{-\totalheight}{\includegraphics[width=2cm]{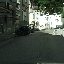}}  \\

  \multirow{1}{*}[-8ex]{\rotatebox{90} {Ground Truth}} & \raisebox{-\totalheight}{\includegraphics[width=2cm]{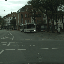}} & \raisebox{-\totalheight}{\includegraphics[width=2cm]{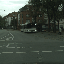}} & \raisebox{-\totalheight}{\includegraphics[width=2cm]{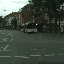}} & \raisebox{-\totalheight}{\includegraphics[width=2cm]{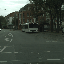}} & \raisebox{-\totalheight}{\includegraphics[width=2cm]{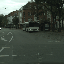}} \\
  & \raisebox{-\totalheight}{\includegraphics[width=2cm]{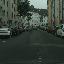}} & \raisebox{-\totalheight}{\includegraphics[width=2cm]{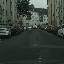}} & \raisebox{-\totalheight}{\includegraphics[width=2cm]{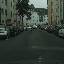}} & \raisebox{-\totalheight}{\includegraphics[width=2cm]{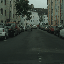}} & \raisebox{-\totalheight}{\includegraphics[width=2cm]{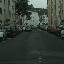}} \\
  \end{tabular}
  \vspace{0.5cm}
  \captionof{figure}{More randomly generated videos based on the Cityscapes dataset.}
  \label{fig:9}
\end{table}

\newpage

\section{Human Evaluation}
\label{appendix:human_evaluation}

\begin{figure}[ht]
    \centering
    \includegraphics[width=.6\textwidth, trim={0.5in 0 2in 0.5in}, clip]{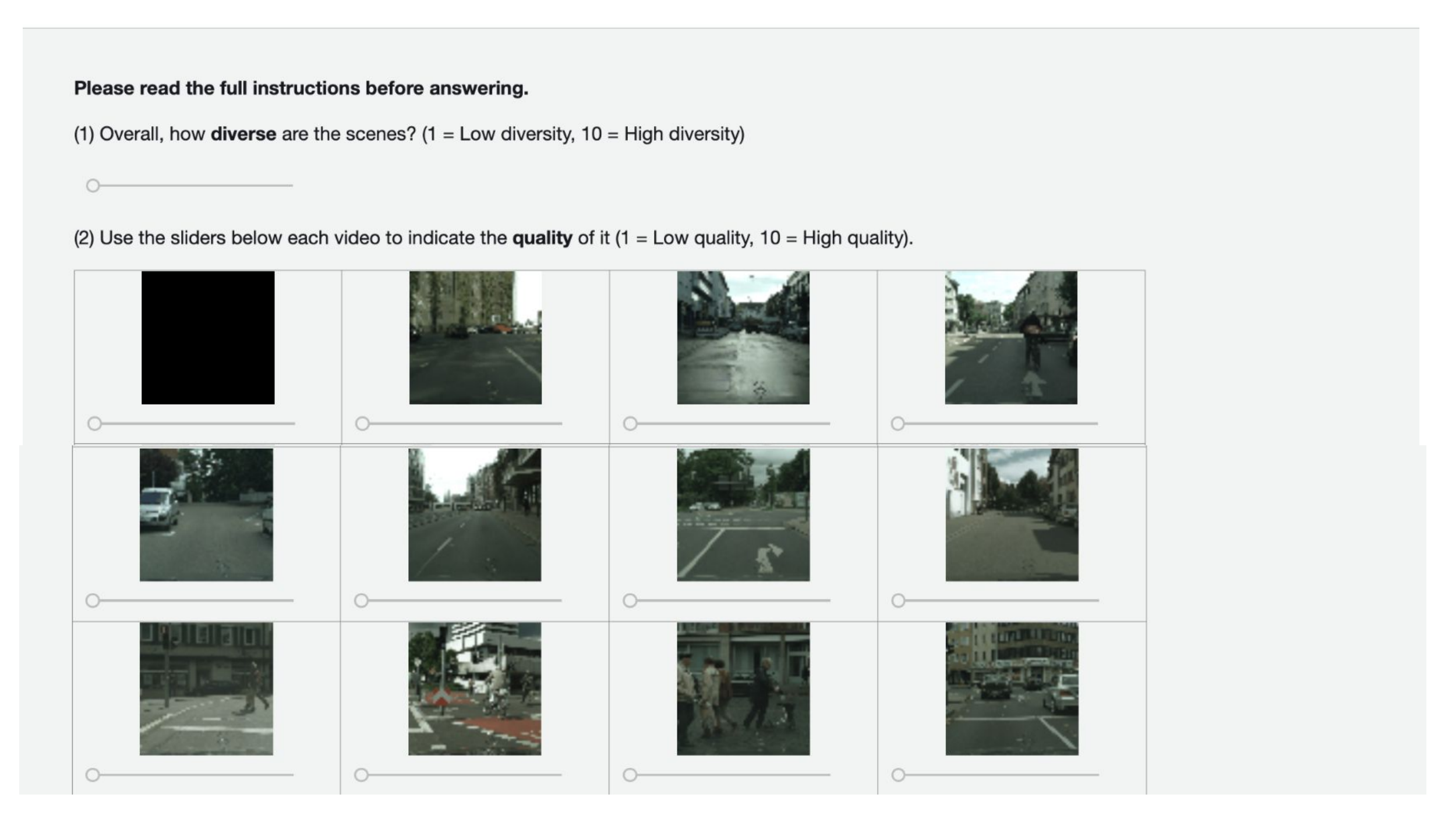}
    \caption{\small{\textbf{Quality and Diversity}. We present to raters 10 videos from the same method and ask them to rate the quality of each video as well as the overall diversity. }}
    \label{fig:mturk_exp1}
\end{figure}

\begin{figure}[ht]
    \centering
    \includegraphics[width=.6\textwidth,  trim={0 0 0 0}, clip]{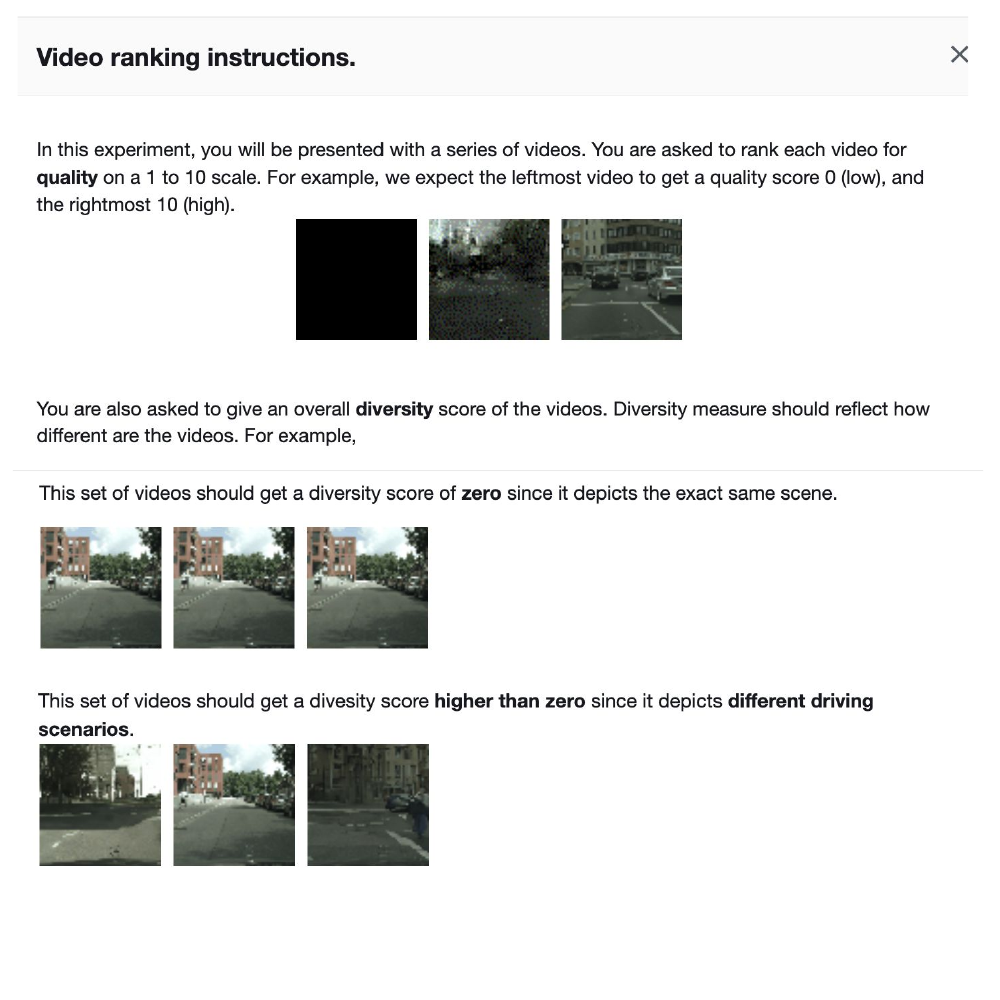}
    \caption{\small{Instructions for the first task.}}
    \label{fig:mturk_exp1_instructions}
\end{figure}

\begin{figure}[ht]
    \centering
    \includegraphics[width=.6\textwidth, trim={0 0 0 0}, clip]{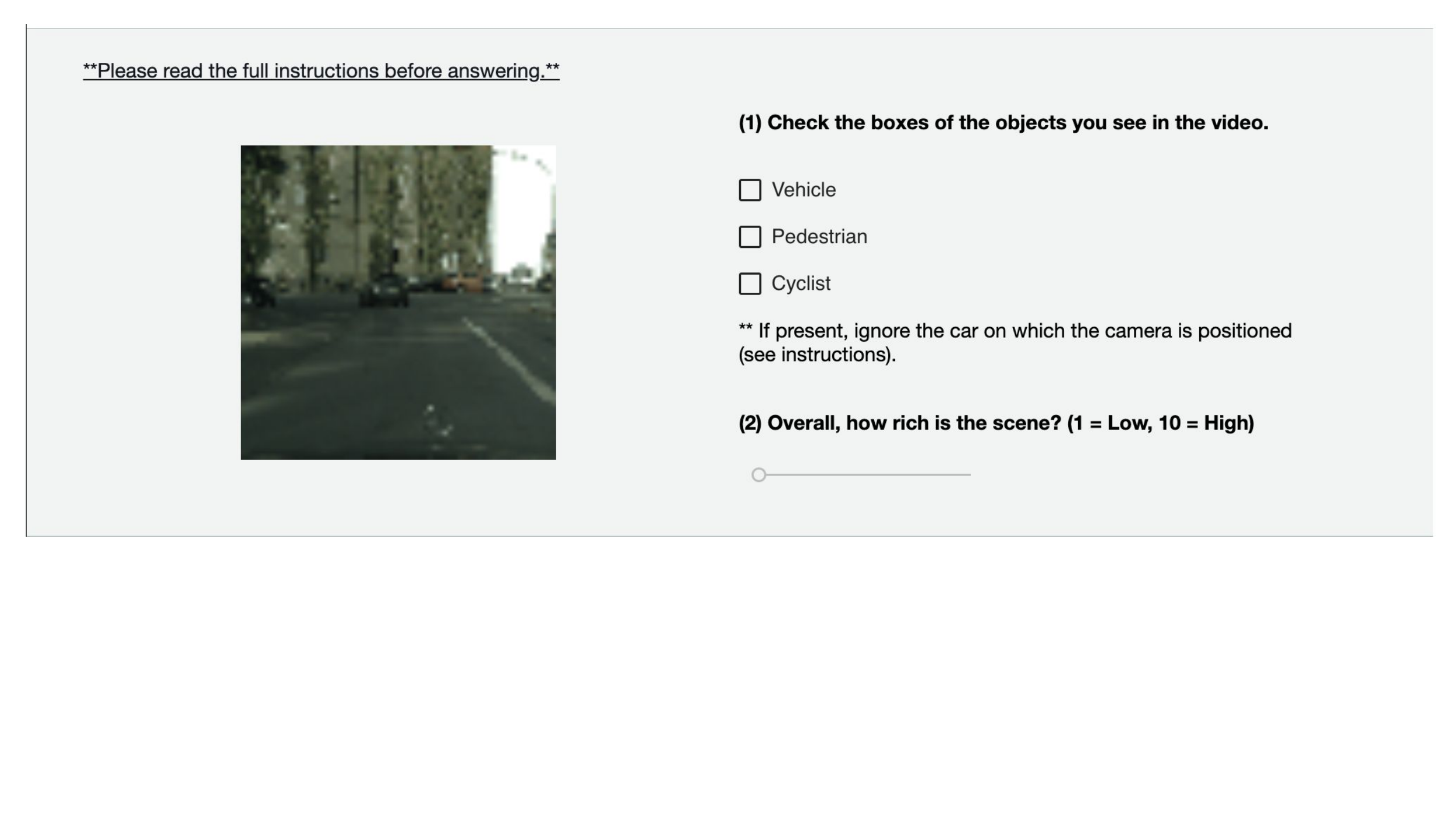} 
    \caption{In a second experiment, we explicitly ask rateres to identify specific objects in the scene: Vehicle, Pedestrian, and Cyclist. }
    \label{fig:mturk_exp2}
\end{figure}

\begin{figure}[ht]
    \centering
    \includegraphics[width=0.6\textwidth, trim={0 0 0 0}, clip]{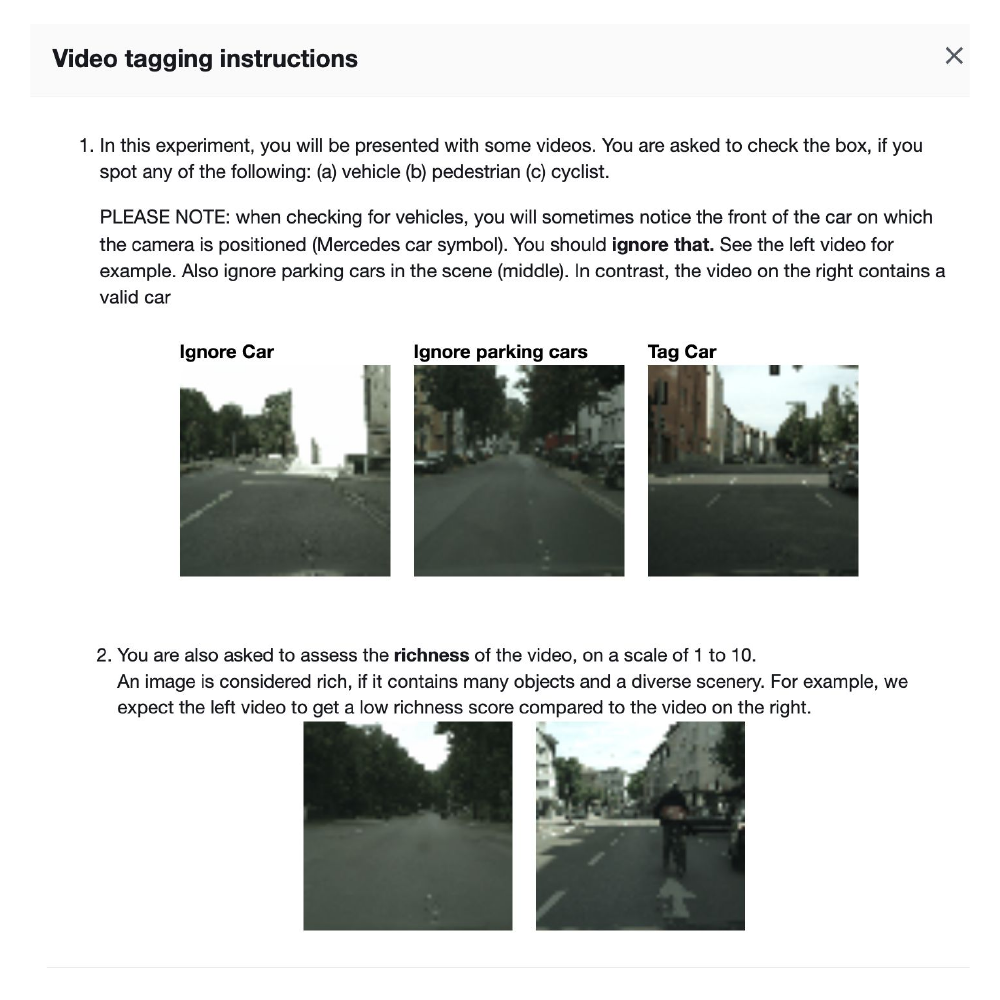}
    \caption{Instructions for the second task.}
    \label{fig:mturk_exp2_instructions}
\end{figure}

The screenshots from the Amazon Mechanical Turk evaluation we conducted are depicted in figures \ref{fig:mturk_exp1}, \ref{fig:mturk_exp1_instructions}, \ref{fig:mturk_exp2} and \ref{fig:mturk_exp2_instructions}.

\end{document}